\documentclass[10pt,twocolumn,letterpaper]{article}

\usepackage[pagenumbers]{cvpr} 
\usepackage[accsupp]{axessibility} 
\usepackage{diagbox}
\usepackage{graphicx}
\usepackage{amsmath}
\usepackage{amssymb}
\usepackage{booktabs}
\usepackage{color}
\usepackage[dvipsnames,table]{xcolor}
\usepackage{adjustbox}
\usepackage{multirow}
\usepackage{pifont}
\definecolor{darkgreen}{RGB}{0,127,0}

\newcommand{\TODONE}[1]{}
\usepackage[pagebackref,breaklinks,colorlinks]{hyperref}
\usepackage[capitalize]{cleveref}
\crefname{section}{Sec.}{Secs.}
\Crefname{section}{Section}{Sections}
\Crefname{table}{Table}{Tables}
\crefname{table}{Tab.}{Tabs.}

\def\cvprPaperID{11460} 
\def\confName{CVPR}
\def\confYear{2022}

\newcommand{\clstoken}[1]{x_{\texttt{cls}}^{\rm #1}}
\newcommand{\dataset}[1]{\mathcal{D}_{#1}}

\begin{document}

\title{Fine-tuning Image Transformers using Learnable Memory}

\author{
\begin{tabular}{c c c c}
Mark Sandler & Andrey Zhmoginov & Max Vladymyrov & Andrew Jackson \\
\multicolumn{4}{c}{Google Inc.} \\
\multicolumn{4}{c}{\{sandler, azhmogin, mxv, jacksona\}@google.com}
\end{tabular}
}

\maketitle

\begin{abstract}
In this paper we propose augmenting Vision Transformer models with learnable memory tokens. 
Our approach allows the model to adapt to new tasks, using few parameters, while optionally preserving its capabilities on previously learned tasks. At each layer we introduce a set of learnable embedding vectors that provide contextual information useful for specific datasets. We call these ``memory tokens''. We
show that augmenting a model with  just a handful of such tokens per layer significantly improves accuracy when compared to conventional head-only fine-tuning, and performs only slightly below the significantly more expensive full fine-tuning. We then propose an attention-masking approach that  enables extension to new downstream tasks,
with a computation reuse. In this setup in addition to being parameters efficient, models can execute both old and new tasks as a part of single inference at a small incremental cost. 
\end{abstract}

\section{Introduction}
\label{sec:intro}
Transformers~\cite{transformers}, originally introduced for sequence problems, such as NLP and speech recognition, have been very successful in advancing state-of-the-art for vision tasks.  ``Vision transformers'' most commonly have been trained on large amounts of data, either in supervised~\cite{vit} or self-supervised~\cite{generative-pretraining}
modes. The resulting model is then ``fine-tuned'' on a downstream task, either in whole or in part. This fine-tuning has been shown to help achieve high performance on a variety of downstream tasks, from classical tasks like image super-resolution~\cite{chen2021pre} and object detection~\cite{beal2020toward} to novel ones, such as Ising model simulations~\cite{kara2021fine}. It has been observed that in order to reach the highest accuracy it is best to fine-tune the entire model on the target task~\cite{vit,generative-pretraining}. However, this is often problematic, since Transformer models can have hundreds of millions or billions of parameters, making maintaining, transmitting and storing a full size  model for  each new task an expensive proposition. Additionally, full fine-tuning is often  sensitive to the choice of the learning rate~\cite{kfortheprice}. In this paper we propose a method where we augment a pre-trained model with learnable tokens that are appended at {\em each} layer of the transformer block.  These extra tokens can be intuitively thought of as a permanent memory that are used as a contextual reference to improve the final prediction.  Another point of analogy is that these tokens can represent concepts at varying levels of abstractions: starting from tokens that run alongside pixel-based  patches, and all the way to the tokens representing final concepts.
\begin{figure}[t] 
  \centering 
   \includegraphics[width=0.95\linewidth]{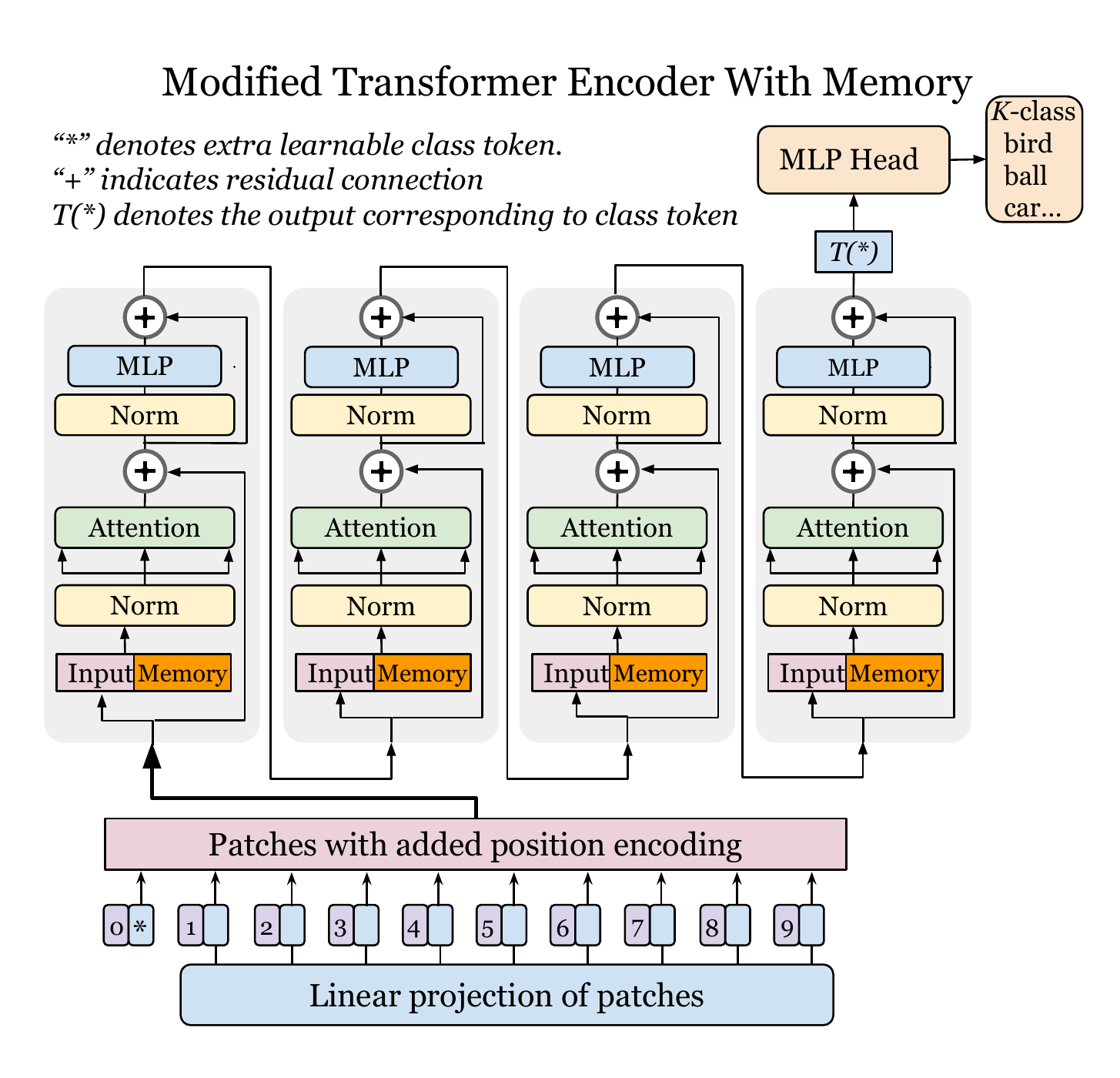}
   \caption{Structure of an encoder with learned memory. The input to each encoder layer  is augmented with learned memory, which is used as an optional source of attention by the embeddings that propagate from previous layers. Memory tokens do not attend to other tokens. For detailed design of individual layer see \cref{fig:memory-model-layer}.\\
   } 
   \label{fig:memory-model} 
\end{figure}

\begin{figure*}[t] 
  \centering 
  \includegraphics[width=0.95\linewidth]{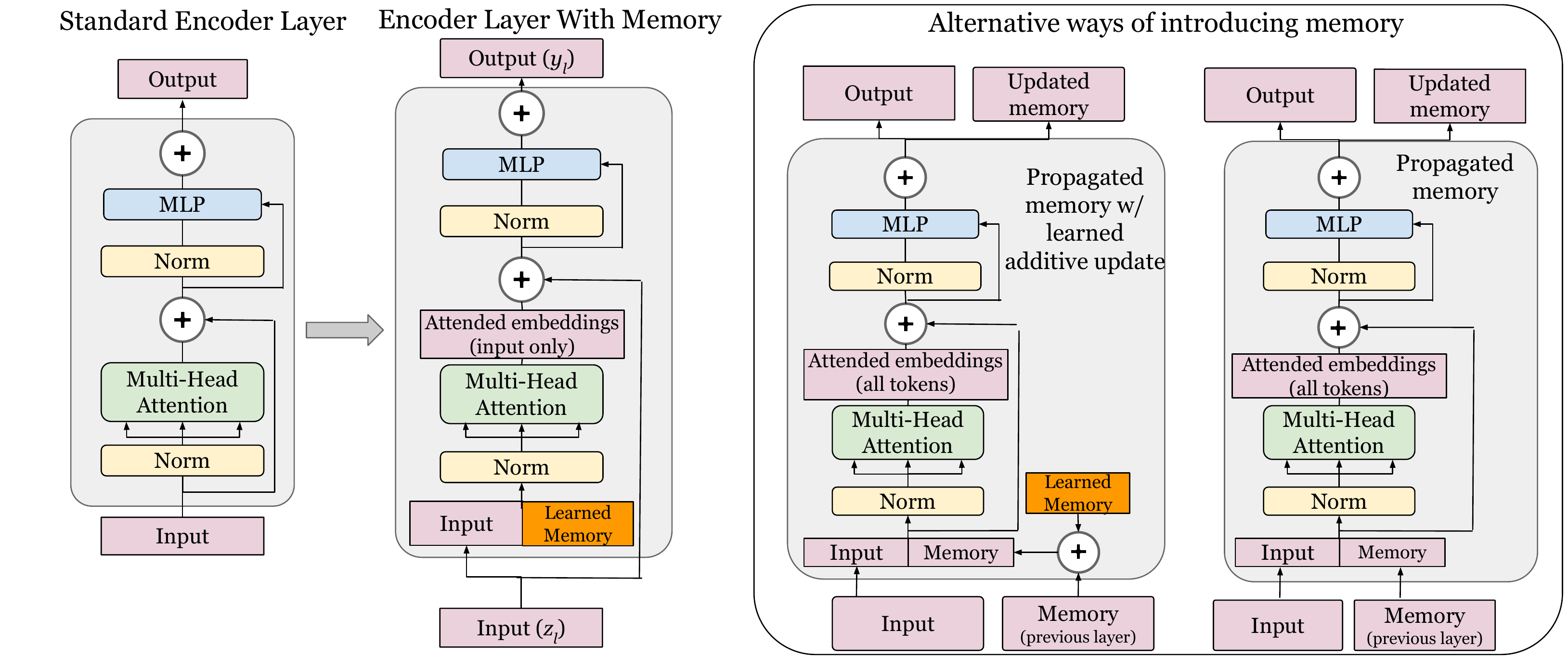}  
   \caption{Detailed design of  encoder layer with memory. The designs on the right are the alternative options that we explore in \cref{sec:experiments}. The rightmost one is similar to the MemTransformer design in~\cite{lowlevelmemory}.
   Here we use rounded boxes to indicate functions and square boxes to indicate input/output to highlight data propagation.} 
   \label{fig:memory-model-layer} 
\end{figure*} 
Our approach allows significant improvements to downstream task accuracy compared to head-only fine-tuning. Further, we believe that our particular architecture design is of independent interest as it provides a novel way of extending a trained model to perform new tasks while retaining the ability of performing old tasks with only incremental compute cost.
This is important in many applications, as the training data for different task often not available at the same time. Naively, by adding more memory one can easily extend a model to new tasks, however, if we try to run such model on an old task we would commonly observe 20\%-40\% percentage drop in accuracy.  We show that with proper attention masking, additional cells can be introduced in a way that doesn't degrade the accuracy of the previous tasks in the presence of new memories,
which is crucial for applications in life-long and continual learning.
We show that the models can be both {\em extended}, where new functionality is added on top of existing ones, as well as {\em concatenated}, where the two (or more) independently trained models can be combined into a single network performing all tasks, with only incremental cost per task. 

The paper is organized as follows. In \cref{sec:model} we describe the method of our memory model. Then in~\cref{sec:masking}  we introduce our attention masking technique that independent sequential task training, without accuracy degradation. Then in \cref{sec:related} we provide an overview of related approaches to memory and  fine-tuning approaches. \cref{sec:experiments} contains our experiments and ablation studies. The last \cref{sec:conclusion} draws some early conclusions and identifies future areas of interest.

\section{Memory model}
\label{sec:model}
\def\RR{\mathbb{R}}
To introduce memory, we build upon the Vision Transformer (ViT) encoder model~\cite{vit}, where an image $x \in \RR^{h\times w\times c}$ is split into a grid of $N$  patches, each of size $P \times P \times c$, which are flattened and fed through the multiple layers of the transformer encoder. For this paper, we only consider classification models, and do not use a decoder. Following~\cite{vit} we use learnable 1D  position embeddings that are added to the input of the first layer. The input to the standard image transformer is defined  as:
$$
\mathbf{z}^{vit}_0 :=  [\clstoken{}, E x_1, \dots,E x_{N}] + E_{pos}
$$
where $E$ is a learnable linear transformation into the encoder embedding space and $E_{pos}$ is a position embedding. Here $\clstoken{}$ is a special learnable token  that is shared for all inputs and the corresponding output value is used as an embedding for the final classification as shown in \cref{fig:memory-model}.  $x_1 \dots x_N$ are flattened image patches. The basic structure of the encoder layer is shown in ~\cref{fig:memory-model-layer}. We refer to~\cite{vit} for the detailed definition of each component. 

Now, to add memory, we concatenate $m$ learnable embeddings $E_{mem}\in \RR^{m \times D}$,
 where $D$ is the dimensionality of the input tokens, as an input to the layer: 
$$
\mathbf{z}^{mem}_0 :=  [\mathbf{z}^{vit}_0; E^0_{mem}]
$$
Thus our transformer receives $N + 1 + m$ tokens in total as an input. This input is then passed through a sequence of encoder-layers as in \cite{transformers, vit}. The architecture of the individual layers exactly matches that of ViT, with a notable exception that we do not propagate updated memory: that is, the output of the self attention module is truncated to the first $N+1$ tokens. Thus the output of the layer $l$, which we denote as $\mathbf{y}_l$, has only $N+1$ tokens.  See the left part of~\cref{fig:memory-model-layer} for a comparison between a standard layer block and our augmented version. 

All consequent layers receive input that follows the same pattern:
$$
\mathbf{z}^{mem}_l = [\mathbf{y}_{l-1}; E^l_{mem}]
$$
where $\mathbf{y}_{l-1}$ is the truncated output of the previous layer. 

In our experiments we explored several different variants of memory models, including ones where memory itself actively attends and propagates to the next layer, however we found that such modifications generally hurt performance. We report some of those results in the ablation study in~\cref{sec:ablation-study}.




\subsection{Fine-tuning with full attention}
The main use of memory that we explore in this paper, 
is applying it to fine-tuning existing models. To do this we introduce randomly-initialized  memory tokens as described above  and perform gradient descent to learn the  memory token, 
the classifier head, and the class token $\clstoken{}$.  While this method gives excellent results, the resulting model can no longer be used to solve its original task, due to change in hidden activations. In the next section we  introduce {\em Attention Masking} that allows to lift this constraint. 
\newcommand{\cls}[1]{\textsc{cls{#1}}}
\newcommand{\mem}[1]{\textsc{mem{#1}}}
\newcommand{\inp}{\textsc{inp}}
\newcommand{\sz}{\footnotesize}
\newcommand{\Y}{\small\cellcolor{green!25}\checkmark}
\newcommand{\X}{\small\cellcolor{red!25}-}
\newcommand{\E}{\small\cellcolor{blue!25}\ding{72}}

\subsection{Attention masking for computation reuse}
\label{sec:masking}

\begin{table}
    \centering
    \scalebox{0.83}{
    \begin{tabular}{c|cccccccccc}
         \backslashbox{\small Q}{\small K}
                    & \inp & \cls{} & $C_1$ & $C_2$ & \multirow{6}{*}{\scriptsize\dots} & $C_k$ & $M_1$ & $M_2$ & \multirow{6}{*}{\scriptsize\dots} & $M_k$ \\  
         \cline{1-5}\cline{7-9}\cline{11-11} 
         \inp       &  \Y  & \Y & \X & \X    & & \X & \X & \X & & \X \\
         \cls{}     & \Y   & \Y & \X & \X    & & \X  & \X & \X & & \X \\
         $C_1$   & \Y   & \Y & \Y & \X & & \X & \Y & \X & & \X \\
         $C_2$   & \Y   & \Y & \E & \Y & & \X & \E & \Y & & \X \\
        \multicolumn{4}{r}{\small{\dots}} & \multicolumn{4}{r}{\small{\dots}}  \\
         $C_k$   & \Y   & \Y   & \E & \E    & & \Y & \E & \E & &  \Y\\
    \end{tabular}
    }
    \caption{Attention mask for model extension and concatenation. Here {\small \begin{tabular}{c} \Y\end{tabular}} indicates that corresponding token type $Q$ (query) attends to token type $K$ (key), {\small \begin{tabular}{c}\X\end{tabular}} that it does not attend, and {\small \begin{tabular}{c}\E\end{tabular}} indicates that attention is used for model extension, but not for concatenation. For brevity, we denoted \cls{-k} as $C_k$ and \mem{-k} as $M_k$ and omitted memory rows since they do not attend to other tokens. See~\cref{sec:masking} for details. 
    }
    \label{tab:attention-mask-preserving-capability-ext}
\end{table}

After fine-tuning the class token of a transformer model on a new dataset, or adding memory, the network performance on the original task generally degrades. This might not be a problem if we are only interested in solving the new task. However it is often desirable to be able to execute both original and new tasks as part of the same inference, with only incremental cost per each new task. A common way to approach this is via multi-task learning where all datasets are present at training time and we learn a universal model with a shared backbone and per-task head  solves all tasks. However, it is not always possible, as in practice data is often owned by different entities, and can not be learned within the same environment or at the same time.  
In this section we describe an extension of our model where  we introduce memory 
a new dataset class token $\clstoken{new}$ and a per-task head  in such a way that the model output for the original dataset token $\clstoken{}$ is preserved, i.e., identical to the output obtained for this dataset token in the original model.
As a result, we can solve multiple tasks for the same input, effectively reusing {\em both} parameters {\em and} computation.
We achieve this by using an attention mask that prevents the original $\clstoken{}$ and input patches from attending to the memory tokens and $\clstoken{new}$ that we introduce for the new dataset. A head for each task is attached to the corresponding dataset token. (see~\cref{tab:attention-mask-preserving-capability-ext}). 

\begin{figure}[t] 
  \centering 
  \includegraphics[width=0.99\linewidth]{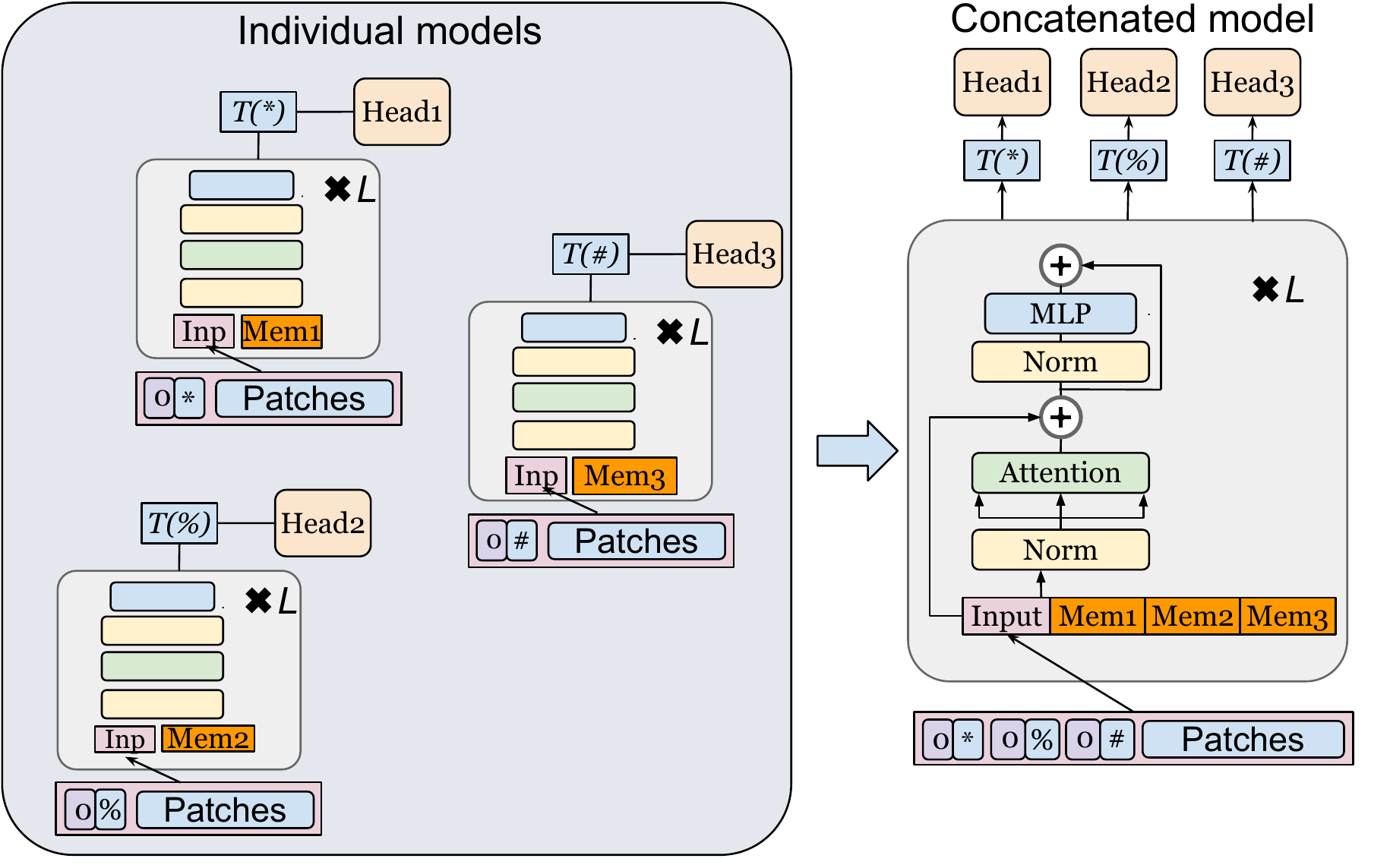}  
   \caption{Concatenating multiple models together. The class tokens are concatenated at the input level,
   while individual models' memory are concatenated at each layer.} 
   \label{fig:model-concatenation} 
\end{figure}

\vspace{-2ex}\paragraph{Model concatenation vs. model extension.}
Attention masking can be used to extend memory as more and more tasks are added. For model extension each new dataset $\dataset{k}$ is trained by fine-tuning its individual memory and the dataset class token $\clstoken{(k)}$ and the attention is constrained to explicitly prevent older datasets from attending to tokens added by following tasks (see ~\cref{tab:attention-mask-preserving-capability-ext}).
Alternatively, if we have two independently trained subtasks\footnote{trained with attention masking preventing $\clstoken{}$ and input patches from attending to $\clstoken{new}$}, we can {\em concatenate} these models into one as shown in ~\cref{fig:model-concatenation}.
In this case the attention mask for the class token $\clstoken{(k)}$ in the concatenated model will be chosen to only attend to $\clstoken{}$ and input patches, and not to other task tokens or memory, thus allowing us to concatenate memory and tokens for models trained independently.
Interestingly, as we discuss in \cref{sec:masking_experiments} we did not see much improvement when extending a model, despite it theoretically being more powerful, as opposed to concatenating them.
However we believe this remains an interesting direction to explore in future work. 

\section{Related work}
\label{sec:related}
Transformers have been the subject of extensive research in the last few years, first in the NLP community~\cite{transformers}. Vision transformers were first introduced in~\cite{vit, deit}, where they achieved impressive results on a series of complex classification tasks. We refer to~\cite{transformers-in-vision-survey,liu2021survey} for an in-depth survey of recent literature. 

There have been a few attempts to couple a learning network with an \emph{external} immutable memory via a learned differentiable module for encoding and retrieval of the information \cite{graves2014neural, weston2014memory, santoro2016meta}. Subsequent work focused on representing memory using a more explicit data structure, such as stacks~\cite{joulin2015inferring}. More recently, in the context of transformers for language modeling, several papers~\cite{guu2020retrieval,fevry2020entities,verga2020facts,lewis2020retrieval,de2021mention} propose to augment a model with access to fixed knowledge base. 


On the other hand, \emph{internal} episodic memory in deep neural networks has been subject of active research since early on, often in the context of recurrent networks. Hopfield networks~\cite{hopfield1982neural} were one of the first attempts to store information within the layers of a neural network. In a recent years, there has been proposed a series of improvements to the energy function of such networks to increase its capacity~\cite{krotov2016dense}, help protect against adversarial attacks~\cite{krotov2018dense} or as a replacement for attention in transformer networks~\cite{ramsauer2020hopfield}.

More generally, different variations of internal recurrent memory networks are actively used in many sequential models, such as in GRU~\cite{GRU} or LSTM~\cite{HochSchm97} architectures. For LSTM, the recurrent cells could be ``enabled'' to preserve or forget a cumulative state as a document was scanned sequentially. Transformers represent a detour where instead of keeping the intermediate state as sequential data is scanned, they introduce a global attention mechanism that enables individual tokens to do random access to the full window of data. As a result, the memory is limited to the size of the window of data, which scales quadratically with respect to the attention window. Several papers attempted to reduce this cost by either introducing sparsity~\cite{child2019generating}, low-rank approximations~\cite{wang2020linformer} or splitting the attention matrix into hierarchical modules connected locally~\cite{beltagy2020longformer}. In a similar vein,  BigBird~\cite{zaheer2020big} and ETC~\cite{ainslie2020etc} models introduce special global tokens that are not part of the input and serve as connector tokens for locally connected segments. More recent work expanded on the use of memory to preserve connections in long sequences to avoid the high computational cost of the full attention matrix~\cite{memorizing-transformers,wu2020memformer}. The main difference from our paper is that the memory considered in earlier works was episodic -- it was preserved only for the duration of the sequence. In the present work the memory is {\em learned} as part of the training process and it does not represent information about any individual samples, but rather is a collective summary representing the entire dataset.

The closest approach to our work is a recent paper by Burtsev \etal~\cite{lowlevelmemory} and that considered a special case of memory where individual memory tokens were added at the first layer alongside the data and were treated as regular inputs and propagated using the same multi-head attention. A similar line of inquiry was also considered in Li \etal~\cite{li2021prefix}, where they considered single task-specific token. All these work was applied to NLP tasks. In contrast, we learn memory that is passed alongside inputs, as a source of attention, {\em at every layer} but it does {\em not} itself attend to any tokens, nor does it propagate to deeper layers. We show that this  improves the performance when compared to only using learned tokens at the input-level layer. Another line of inquiry that uses a similar approach is a recent work on data-efficient transformer training by Touvron \etal~\cite{deit}. There, 
authors train an additional class token for distillation purposes only. Using a dedicated token for a new task can be seen as a special memory based approach. This was further refined in the follow-up work by the same authors in~\cite{touvron2021going}, where they moved the class token into later layers of the architecture.  The class token is then used as a read-out token. This can be seen as a hybrid  of propagated-memory and finetuning the class token approach.

An alternative way of viewing the present work can be seen as modifying the individual layers during fine-tuning to redirect their attention in novel ways. Such approaches were explored in ~\cite{houlsby2019parameter,pfeiffer2020adapterhub,hyperformers}. There, a set of special layers called adapters are introduced for each of the fine-tuned tasks. During fine-tuning, only those layers and/or normalization parameters~\cite{kfortheprice,yu2018slimmable} are updated.  Such approaches are more complex because they introduce additional structures and  in contrast with our approach, they  do not allow non-destructive fine-tuning.

%
\def\ViT{ViT}
\def\cifar{CIFAR-100}
\def\inat{i-Naturalist}
\def\places{Places-365}
\def\imagenet{Imagenet-1K}
\def\sun{SUN-397}
\section{Experiments}
\begin{figure*}
    \centering
    \includegraphics[width=0.9\textwidth]{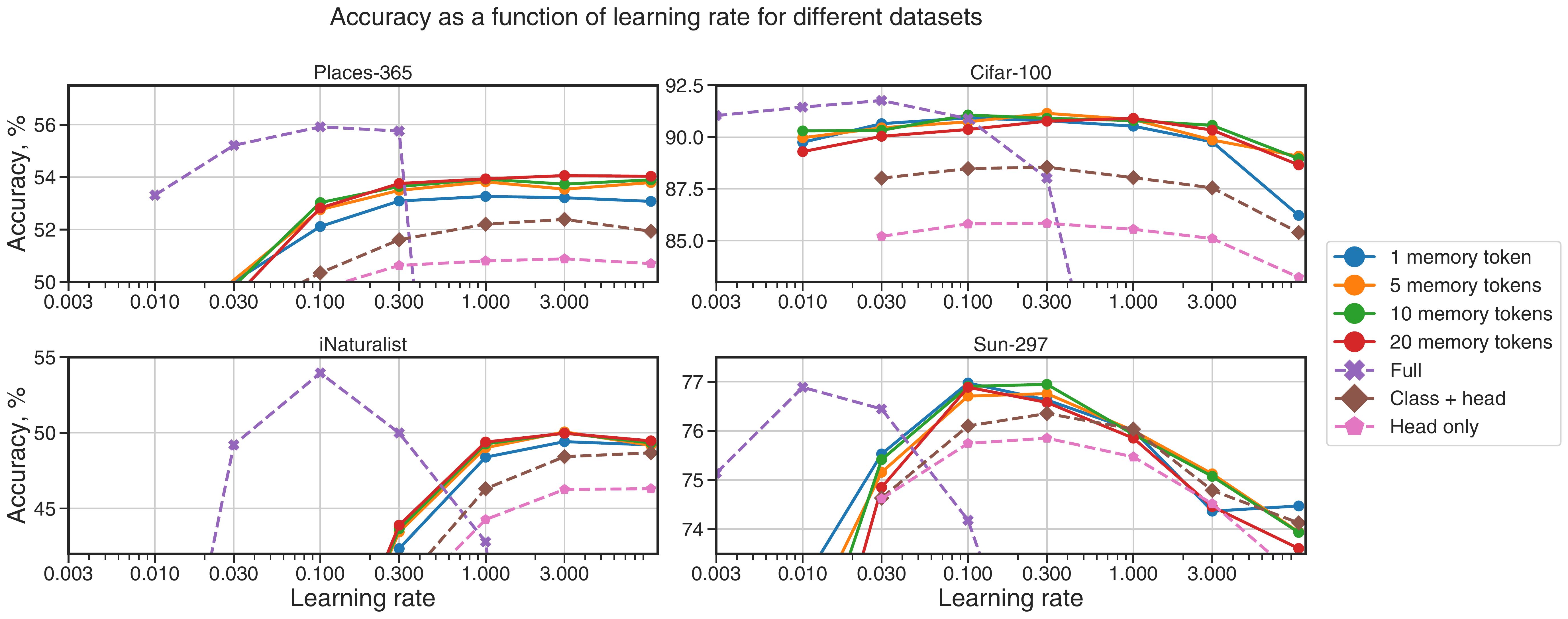}
        \caption{Performance of memory-augmented transformers for different datasets and comparison with baseline fine-tuning methods.
        \TODONE{nit: 25\% of space in Places and CIFAR and 50\% of space in Sun is empty. Consider changing the ylim of Places to [0.475, 0.575] for Places, [0.825, 0.925] for Cifar and [0.725,0.775] for Sun. Also, why lr=0.03 for Sun for Full has 2 points? Also, y-axis says percent, but the scale is a probability}
        }
    \label{fig:fine-tuning-regimes}
\end{figure*}
\begin{table*}[t]
    \centering
    \begin{tabular}{|r||c|cc|cccc|}
\hline
                & Full & Head Only & Head + Class & 1 cells & 5 cells & 10 cells & 20 cells\\
\hline
        Sun-297 & 76.9 & 75.9 & 76.4 & \bf{77.0} & 76.8 & 76.9 & 76.9 \\
    iNaturalist & 54.0 & 46.3 & 48.7 & 49.4 & \bf{50.1} & 50.0 & 50.0 \\
      Cifar-100 & 91.8 & 85.8 & 88.6 & 90.9 & 91.0 & \bf{91.1} & 90.9 \\
     Places-365 & 55.9 & 50.9 & 52.4 & 53.3 & 53.8 & 53.9 & \bf{54.1} \\
\hline
\end{tabular}

    \caption{Accuracy for different datasets using the optimal learning rate for each fine-tuning regime.}
    \label{tab:main_table}
\end{table*}
\begin{table}[t]
    \centering
\begin{tabular}{|r||c|cc|c|}
\hline
                & Full & Head & Head + \cls &  5 cells\\
\hline
        \ViT-B/32 & 55.9 & 50.9 & 52.4 & \bf{53.8} \\
        \ViT-B/16  & 58.0 & 52.3 & 53.8 & \bf{55.4}  \\
        \ViT-L/32  & 56.6 & 52.4 & 49.7 & \bf{54.8}  \\
\hline
\end{tabular}   
\caption{Accuracy for \places{} dataset using  optimal learning rate for different \ViT{} architectures.}
    \label{tab:architectures}
\end{table}
\begin{table}[t]
    \centering
\begin{tabular}{|l||c|c|}
\hline
                  & Accuracy& Params (\ViT-B/16)\\
\hline
        Memory    & 50.1 & 768 * 12 * m $\approx $ 46K   \\
        Adapters~\cite{houlsby2019parameter}  & 49.9 & 3078 * 5 * 4 * m $\approx$ 737K   \\
        Combined  & 50.3 & 46K + 737K $ \approx$ 783K   \\        
\hline
\end{tabular}   
\caption{Performance for \inat{} for adapters~\cite{houlsby2019parameter} with bottleneck of size 5, vs learnable memory with 5 cells. (ours). Parameter counts exclude the size of the head which is the same for all methods}
    \label{tab:methods}
\end{table}
\label{sec:experiments}
\subsection{Datasets and training setup}
For all our experiments, with the exception of \cref{tab:architectures} we use the \ViT-B/32 base transformer model from~\cite{vit}. We use the pre-trained model released in \cite{vit}, which was trained on Imagenet-21K~\cite{imagenet21k}. This model has about 80M parameters. We use a cosine learning rate schedule~\cite{cosine-lr} and  followed the setup of fine-tuning in ~\cite{vit}, to generate our baselines.  We used batch size 512, and all fine-tuning runs were run for 20000 steps. We note that shorter runs generally reached slightly worse results but preserved the relative relationship between different fine-tuning setups. We used SGD with Momentum with gradient clipping, though we did not see much benefit with or without it. We used a 5-step linear rate warmup.

Following the standard practice we used standard inception preprocessing~\cite{inception} for all datasets except \cifar, where we used random clipping. For the latter we only used random flips.  The memory was initialized using ${\cal N}(0, 0.02)$,  following the initialization of other variables in Vision Transformers in ~\cite{vit}. 
We evaluate the performance of our approach on the following datasets
\cifar~\cite{cifar100}, \inat~\cite{inat17}, \places~\cite{zhou2017places} and \sun~\cite{sun}. In all cases we used 95\% of the training split for training, the remaining 5\% as a dataset for early stopping and hyper-parameter selection, and we used the validation split to report final accuracy.


\subsubsection{Baseline fine-tuning experiments}
We compare our methods with the following fine-tuning modes. 
\vspace{-2ex}\paragraph{Full fine-tuning} the entire model is fine-tuned. This is the most expensive and thus often the least practical regime if multiple downstream tasks are present, because each task requires a full model. It is also prone to overfitting and is sensitive to the learning rate. For example, we found that often the optimal learning rate for other regimes resulted in full fine-tuning producing dramatically sub-optimal results, as can be seen in \cref{fig:fine-tuning-regimes} 
\vspace{-2ex}\paragraph{Head-only fine-tuning} as might be expected, only tunes the head of the classifier. This approach essentially fully reuses the embedding learned by a pretrained model. One important advantage of this mode is that it allows reuse of both the {\em parameters} and the {\em compute}, because the computation of embeddings is shared across tasks. 
\vspace{-2ex}\paragraph{Head + Class token} where we fine-tune the head and the class token. Note that this regime, while using only marginally more parameters, does not allow any significant computation reuse, since an input tokens attention (and embedding) will change as they attend to a new class-token. We note that this method can be seen as a precursor to memory training, since tuning the class token allows us to learn a representation of a dataset. 

\subsubsection{Memory fine-tuning}
Our main experiments use the following two regimes.
\vspace{-2ex}\paragraph{Memory + head + class token} We refer to this as ``memory'' fine-tuning. This is our method. We conduct experiments on architectures containing 1, 2, 5, 10 and 20 cells. We did not see improvement when using more than 20 cells per layer.
In fact most of the benefit is realized (see \cref{tab:main_table}),  when we use 5 cells for all our datasets, thus we conjecture that using 5 cells provides good universal size that can be used without any additional hyper-parameter search. 

In \cref{fig:fine-tuning-regimes} we show performance of learned memory vs other fine-tuning methods for different learning rate. As can be seen, especially fine-tuning performs best as much lower learning rates. Thus in all our experiments we used the best learning rate for each regime.

\vspace{-2ex}\paragraph{Memory with attention mask} This is the regime which combines the benefit from {\em head-only} fine-tuning of allowing computation reuse, while achieving accuracy above {\em Head + Class} token, and 
slightly below the full-attention fine-tuning. The advantage  of this regime is that models can be independently fine-tuned on multiple tasks, and combined in a single model that reuses most of the compute. 

The number of variables used by our different approaches is shown in~\cref{tab:parameters_cost}.

The rest of the section is organized as split into three parts. In \cref{sec:transfer} we discuss  our results  on standard transfer learning tasks, and in~ \cref{sec:masking_experiments} we discuss the results on attention-masking based  fine-tuning. 
\begin{table}
    \centering
    \begin{tabular}{|l|l|c|c|c|c|}
    \hline
      & Full &  Head only  &  Class  & Mem (5) \\     
      
     \hline
      Params & 80M  &  $768 \times k$   & 768    & $768 \times 12 \times 5$ \\    
      Flops &$ \approx $1G    &  0                & 0      & $\approx$ 25M \\
     \hline    
    \end{tabular}
    \caption{Incremental cost on number of parameters  and FLOPS for \ViT-32-B for each fine-tuning regime. Class and head fine-tuning do not incur any extra computation cost as the architecture stays unchanged.  Here $k$ denotes number of classes, and we assume 5 memory tokens per layer.}
    \label{tab:parameters_cost}
\end{table}

\definecolor{mplblue}{rgb}{0., 0.5, 0.91}

\begin{figure*}[t]
    \centering
    \includegraphics[width=0.9\textwidth]{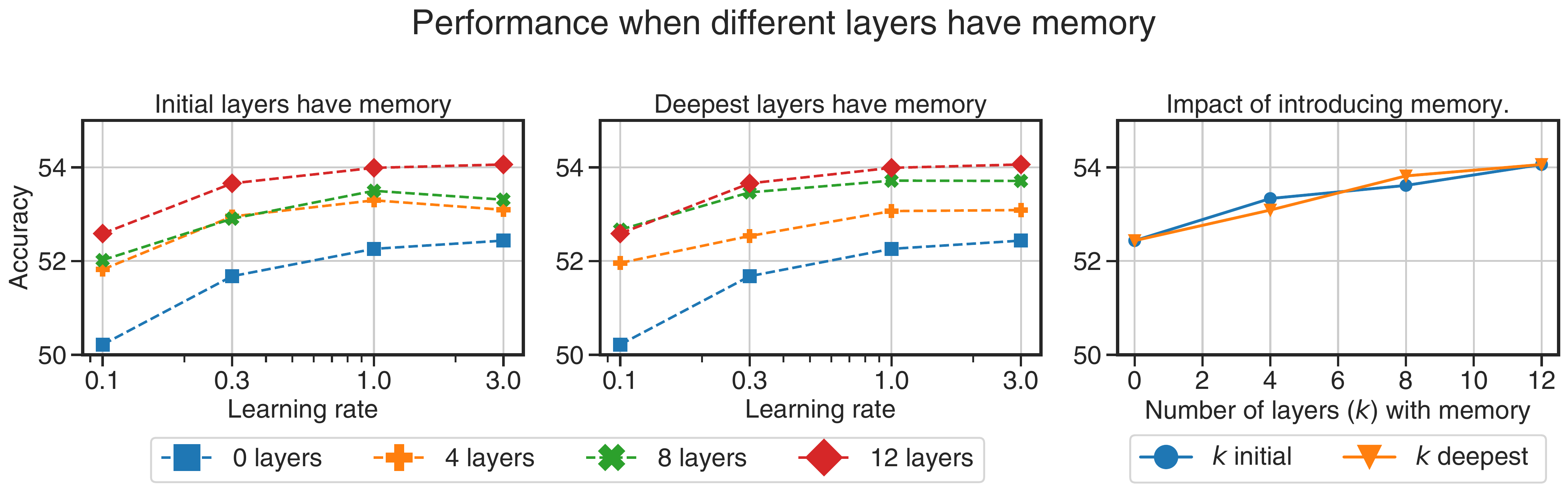}
    \caption{Introducing memory in different layers. All results are for \places~ dataset. The left chart shows the impact of serially introducing memory in the first 0, 4, 8 and 12 layers, respectively the left chart shows the effect of introducing memory in the deepest layers first. The right-most chart shows performance for best optimal learning rate for each count $k$. 
    }
    \label{fig:memory_at_different_layers}
\end{figure*}
\begin{figure}
    \centering
    \includegraphics[width=0.45\textwidth]{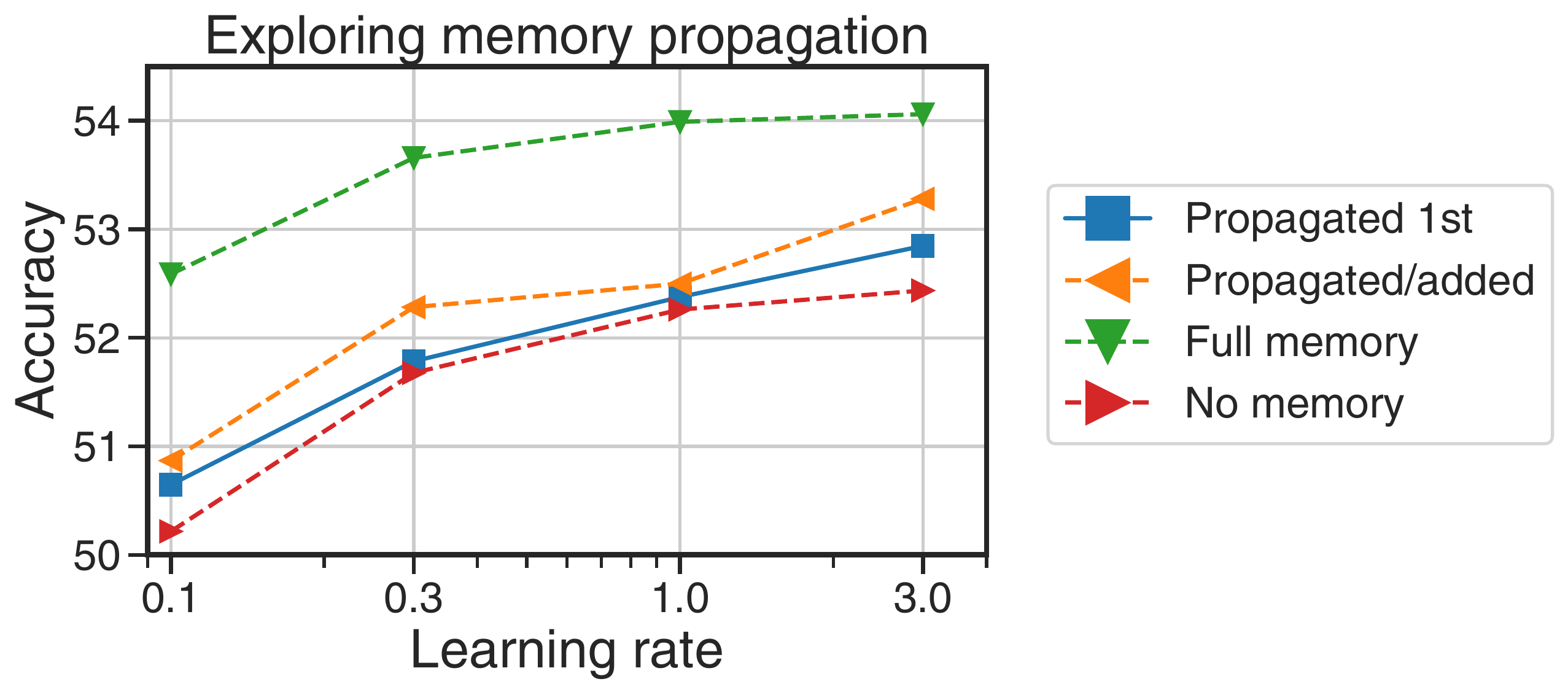}
    \caption{Performance of different  memory architectures. All results are for \places~ dataset.``Full memory"" is our approach with learnable memory at each layer . ``Propagated 1st" propagates memory introduced at the first layer, as described in \cite{lowlevelmemory}. ``Propagated/added" propagates memory from previous layer and adds new at each layer. See \cref{fig:memory-model-layer} and \cref{sec:ablation-study} for more detail. }
    \label{fig:different_memory_types}
\end{figure}

\begin{figure}
    \centering
    \includegraphics[width=0.45\textwidth]{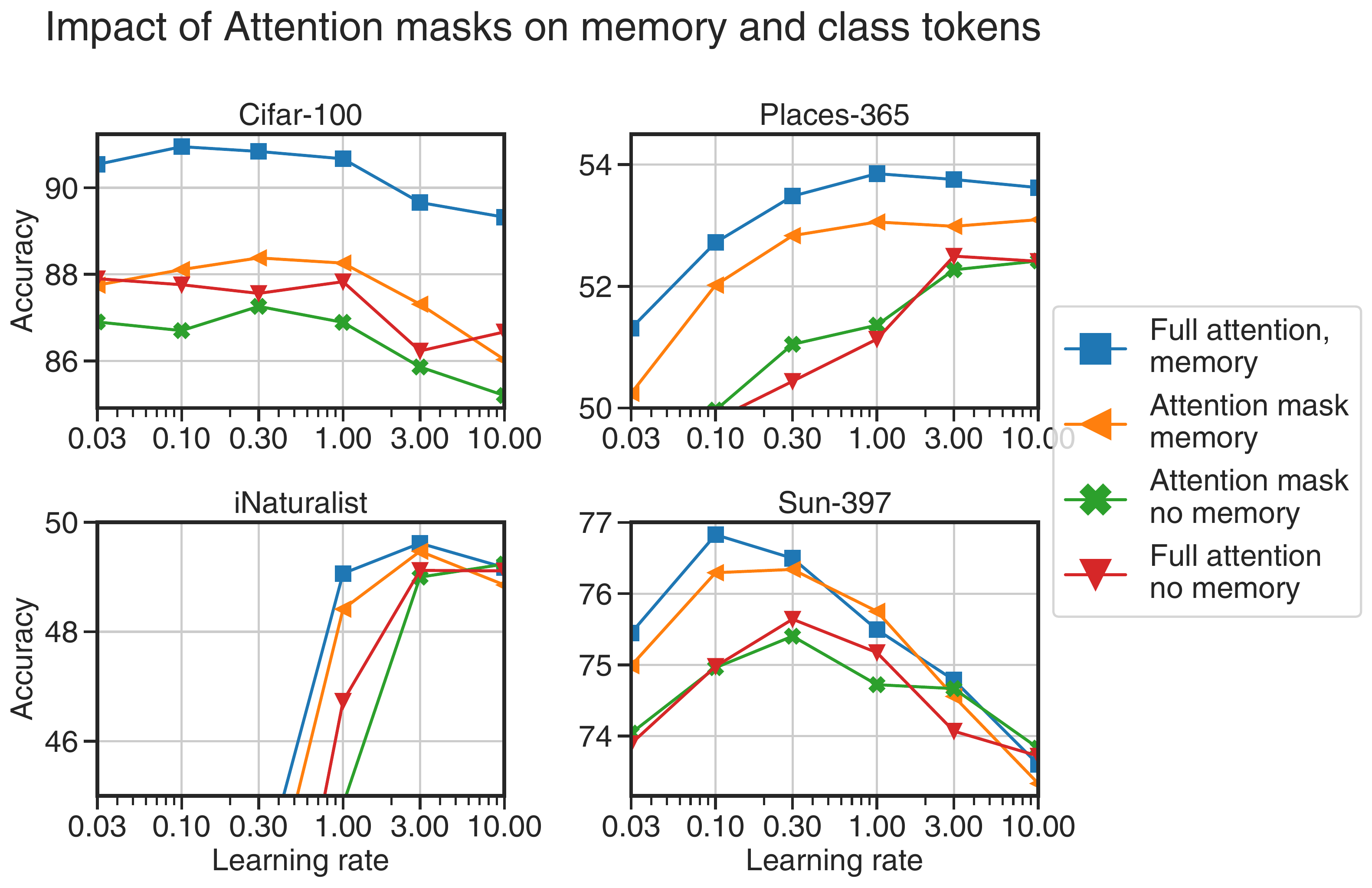}
    \caption{Performance of fine-tuning with attention mask. The top line ($\color{mplblue}\blacksquare$) shows the baseline with full attention. 
    The second ($\color{orange}\blacktriangleleft$) shows attention-mask with memory. The last two are class+head fine-tune with full attention ($\color{red}\blacktriangledown$) and with attention mask~($\color{darkgreen}{\boldsymbol{\mathbf{\times}}}$).
   }
    \label{fig:non-destructive}
\end{figure}

\subsection{Performance of memory for transfer learning}
\label{sec:transfer}
In \cref{fig:fine-tuning-regimes} we show our results for fine-tuning with memory versus the baseline regimes as a function of learning rate. Remarkably,
full fine-tuning is much more sensitive to the learning rate. The performance dropped dramatically for a large learning rate, while memory and
head-only fine-tuning achieve better performance with higher learning rates. As can be seen from the same figure,  models with memory  outperform all other
fine-tuning regimes.  

In  \cref{tab:main_table} we show the comparison where we picked the best learning rate for each fine-tuning regime and for different numbers of memory tokens. 
As can be seen, for most datasets the highest performance is achieved when the number of memory tokens is between 5 and 20. We hypothesize that such a small difference is caused by the increased potential for over-fitting as more memory is introduced. In the supplemental materials we show that training accuracy increases steadily as the number of memory tokens increases.
For Cifar-100 and Sun-297, the performance of memory-fine-tuning nearly reaches the performance of full fine-tuning, while using <1/100 of the number of variables. 

\subsection{Ablation Study}
\label{sec:ablation-study}
We now turn our attention to exploring which components of our approach are important,
as well as showing some alternative approaches. First of all, we consider
introducing memory at the first layer only and propagating through the entire transformer as shown in \cref{fig:memory-model-layer}. 
This can be as seen as introducing additional class tokens as a source of attention for other tokens in the inputs. 
This approach is similar to that introduced in~\cite{lowlevelmemory} for NLP tasks. While we see small improvement, as seen in 
\cref{fig:different_memory_types}, it is minimal compared to improvements from using full memory. 
A variant of this approach that can be seen as hybrid of propagating and learned
memory, is when the memory is both propagated and then learned memory is added. This method worked  better, but overall it appeared 
that using learnable memory works best without propagation. 

Another question we consider is which layers are important for memory introduction. For that we introduce $k$ layers of memory starting from the 
earliest layers (close to input), and compare it against memory introduced at deepest layers. As can be seen in \cref{fig:memory_at_different_layers} the performance gain is mostly gradual. It appears that adding memory to the initial layers has less performance benefit than adding them to the last 4 layers, but it appears that all layers benefit from adding memory.


\subsection{Attention masking}
\label{sec:masking_experiments}
\begin{table*}
    \begin{center}
    \begin{tabular}{|l|c|c|c|c|c|}
        \hline
        Fine-tuning   & Comp. Reuse &  \cifar & \places & \sun & \inat \\
        \hline
        {\em Class, no memory} & N & $88.6$ & $52.4$ & $76.4$ & $48.7$ \\
        {\em Class, 10 memory tokens, full attention} &N & $\bf{91.1}$ & $\bf{53.9}$ & $\bf{76.9}$ & $\bf{50.0}$ \\
        \hline
        {\em Head-only, no memory}      & Y  & $85.8$ & $50.9$ & $75.9$ & $46.3$  \\
        {\em Class, masking, no memory} & Y  & $87.2$ & $52.5$ & $75.8$ & $49.0$ \\
        {\em Class, 10 memory tokens, masking} &  Y & $\bf{88.7}$ & $\bf{53.3}$ & $\bf{76.4}$ & $\bf{49.5}$ \\
        \hline
        \hline
        {\em Full attention, 4 models combined} & -  &$71.5$ & $47.6$ & $67.8$ & $17.3$ \\
        \hline
    \end{tabular}
    \end{center}
    \caption{
		Test accuracies of models fine-tuned from a pre-trained \ViT-B/32 model on different datasets.
		All models included an additional separate class token.
		We consider models trained with full attention and models trained with attention-masking.
		The models with masking can be combined with each other without any accuracy degradation, while also retaining ability to perform original  Imagenet-21K predictions as part fo the same computation.
		To demonstrate {\em interference} of models trained without attention masking, in a separate experiment (last row), we concatenated all models fine-tuned with full attention and memory, and then evaluated different dataset heads.
	}
\label{tab:non-destructive}
\end{table*}

In this section we study attention masking for model fine-tuning as discussed in ~\cref{sec:masking}.
First we verify that performance degrades if we simply concatenate memory and class tokens fine-tuned independently without attention masking.
We start by independently fine-tuning 4 separate models with dataset-specific class tokens and memory on \sun, \places, \inat{} and \cifar.
All four models are fine-tuned from the same pre-trained \ViT-B/32 model.
Starting with a model trained on \inat, we then progressively add class tokens and memory from models trained on \places, \cifar{} and \sun{} in that order.
As a result, the original \inat{} accuracy evaluated using a corresponding class token degrades from $50.0$ to $28.1$ (after adding \places), $27.8$ (after adding \cifar) and finally to $17.3$ (after adding \sun).
In other words, naive concatenation of class tokens and memory pre-trained without attention masking, results in destructive interference between the elements of the sequence, hurting the performance of individual model heads.

Alternatively, we could instead fine-tune individual models with attention masking and concatenate the individual class tokens and memory tokens with a mask that would prevent them from interfering with each other (see~\cref{sec:masking}), retaining the original accuracy of individual models.
Figure~\ref{tab:non-destructive} shows test accuracy for models fine-tuned on different datasets with attention masking.
Our results, summarized in~\cref{tab:non-destructive}, show  attention mask produces models that are less accurate compared to their full-attention counterparts, but there is still an advantage of using additional memory tokens compared to models without them and this approach to training allows us to assemble multiple models into a single model sharing most of the necessary computation while still retaining Imagenet-21K head predictions.

Finally, we conducted experiments where we {\em extended} the original \ViT-B/32 model by adding dataset-specific class tokens and memory for each new consecutive dataset. 
In this setup, as discussed in \cref{sec:masking}, the attention mask restricted old class tokens and memory from attending to tokens added later for new datasets, but new tokens could attend to all existing tokens. As in all our experiments, we started with \ViT-B/32 model and fine-tuned it on two datasets in one of two ways, either: (a) fine-tuned it first on \sun{} and then on \places, or (b) fine-tuned in on \places{} and then on \sun. We did not see any improvement from the case when  \sun{} and \places{} were trained independently and concatenated post-training. We hypothesize that in order to see advantage in model extension, the intermediate datasets need to provide a significant amount of relevant samples compared to both the pre-training dataset and the new final dataset. 

\section{Conclusions, Limitations and Future work}
\label{sec:conclusion}
We proposed a novel method of incorporating memory into transformer models. Our results show that memory improves the performance of a fine-tuned model significantly
and allows training models for multi-task scenarios without compromising on performance. The memory tokens are shown to be an important part of the attention model. We believe they could be a critical ingredient for lifetime and continuous learning.  There are several directions that are subject to future exploration. There are some limitations in our work that will be a subject of future work. 
\vspace{-3ex}\paragraph{Scaling memory. } We observed that as the number of memory tokens increases, there appears to be little benefit from using more than 10 tokens. One potential direction is to explore limiting the attention to the Top-K memory tokens. This would reduce the amount of background noise from irrelevant tokens. 
    
\vspace{-3ex}\paragraph{Combining episodic memory with learnable memory.} Episodic memory provides a way to memorize very long sequences, without remembering everything. This type of memory is typically used during inference and not learned directly. Unifying permanent and episodic memory within the same framework is an interesting extension.
\vspace{-3ex}\paragraph{Incremental Model extension.} We showed that multiple attention masking models can be trained independently and concatenated into one model with efficient computation reuse while outperforming the baseline. However, showing that models trained {\em sequentially} can benefit from intermediate memory accumulation is important for curriculum learning and is a subject of future work. 
    
\vspace{-3ex}\paragraph{Novel domains and architectures} An important next step is to introduce memory into other transformer models, and to apply our method to different domains, such as NLP. 

{\small
\bibliographystyle{ieee_fullname}
\bibliography{main}
}
\appendix
\section{Dependence of training accuracy on  number of memory tokens }
\begin{figure*}[t]
    \centering
    \includegraphics[width=0.95\textwidth]{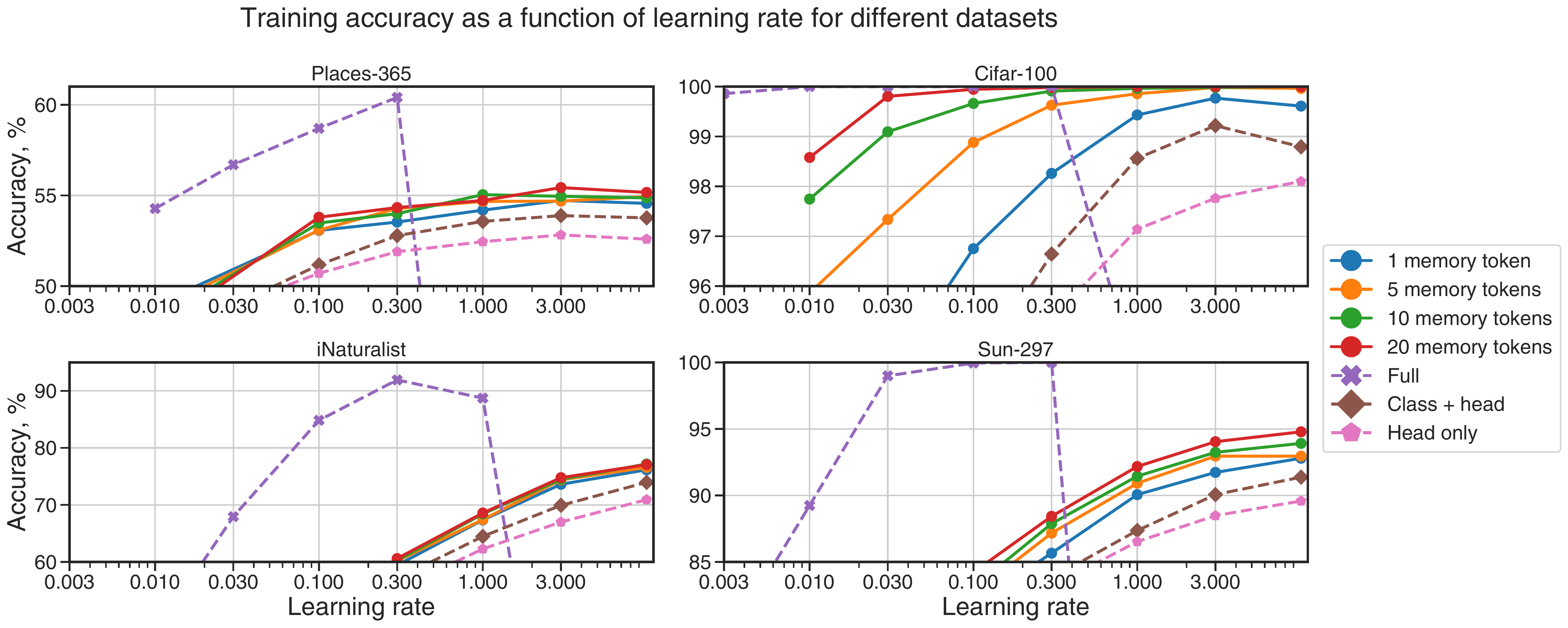}
        \caption{Performance of memory-augmented transformers for different datasets and comparison with baseline fine-tuning methods.      
        }
    \label{fig:fine-tuning-regimes-train}
\end{figure*}
As we discussed in the experiments sections, the test data performance as a function of number of memory tokens appear not to improve
much.  We hypothesize that the reason for that might be the growing generalization gap. Specifically on~\cref{fig:fine-tuning-regimes-train}
we show the training performance of memory-augmented transformers. As can be clearly seen, there the performance improves monotonically as the number of memory token grow.

\section{Exploring memory attention}
\begin{figure*}
    \centering
    \includegraphics[width=0.95\textwidth]{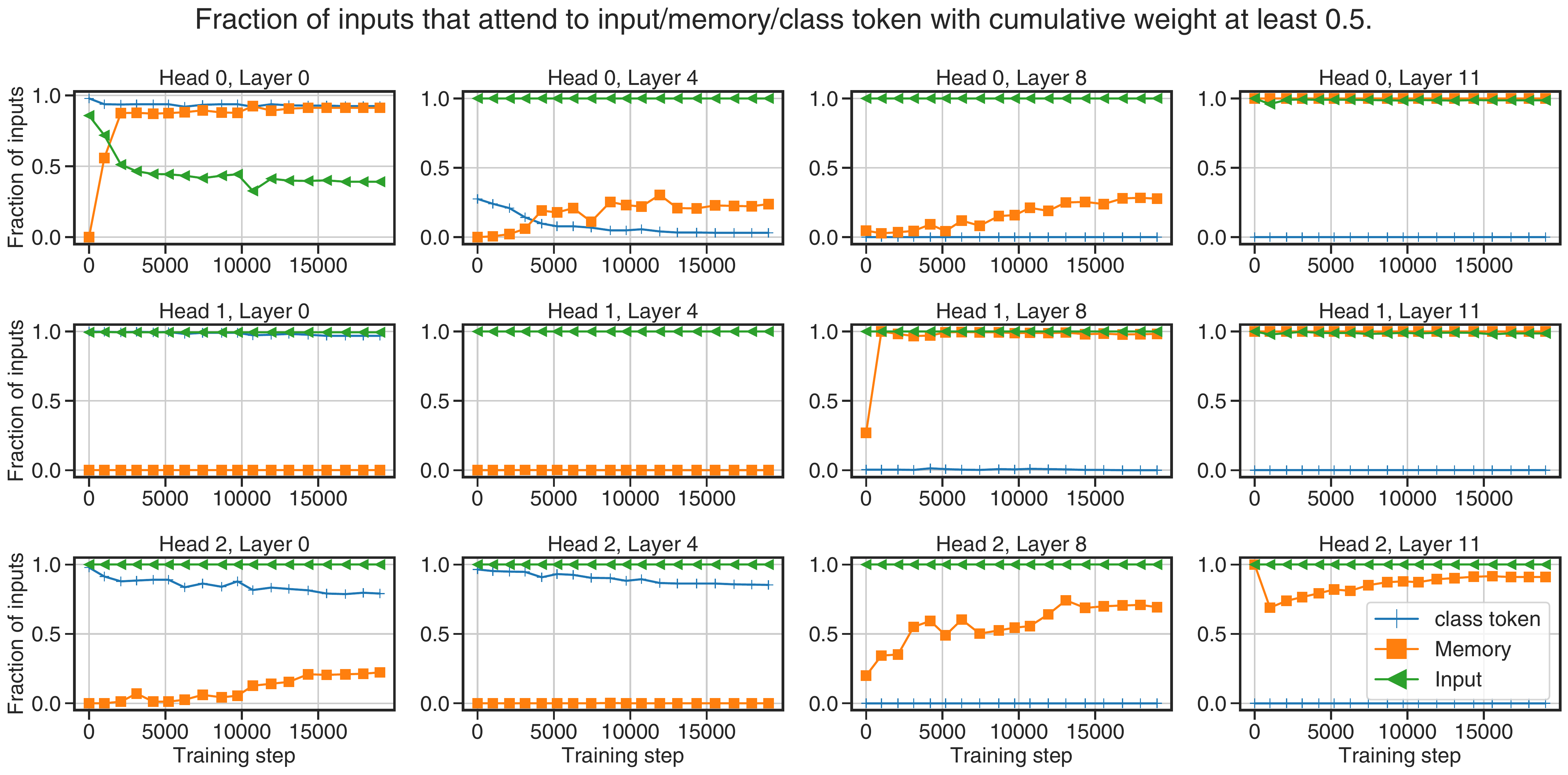}
    \caption{Attention of input tokens to different types of other tokens for individual heads and layers. We only include 3 heads (0, 1,  and 2) and 4 layers spread uniformly over architecture. Here we calculate what fraction of samples have at least one token, that attends with weight at least 0.5  to  (a) memory ($\color{orange}{\blacksquare}$) (b) class token ($\color{mplblue}+$) and (c) self-attention ($\color{darkgreen}{\blacktriangleleft}$). Remarkably,   we see a significant variability for different heads. }
    \label{fig:input_token_attention}
\end{figure*}

In this section we explore how the attention to memory changes as training progresses. In  \cref{fig:input_token_attention} we show how the attention changes over training trajectory. Specifically, we measure the fraction of input samples in the validation subset, that have at least one input token, have cumulative attention to memory of at least 0.5 at different layers. We used the first 3 heads and measured the attention at 4 different layers in the beginning, in the middle and at the top of the network.  As we can see, the general pattern is that attention to memory tends to increase as learning progresses.

\end{document}